\documentclass[conference]{IEEEtran}

\usepackage{amssymb}
\usepackage{amsthm}

\usepackage[cmex10]{amsmath}

\usepackage[noadjust]{cite}

\usepackage{fixltx2e}

\begin{document}

\title{Evolution and the structure of learning agents}

\author{\IEEEauthorblockN{Alok Raj}
\IEEEauthorblockA{Cisco Systems, Inc.\\
San Jose, California 95134--1706\\
Email: ralok@cisco.com}}

\maketitle

\begin{abstract}
This paper presents the thesis that all learning agents of finite information size are limited by their informational structure in what goals they can efficiently learn to achieve in a complex environment. Evolutionary change is critical for creating the required structure for all learning agents in any complex environment.
The thesis implies that there is no efficient universal learning algorithm. An agent can go past the learning limits imposed by its structure only by slow evolutionary change or blind search which in a very complex environment can only give an agent an inefficient universal learning capability that can work only in evolutionary timescales or improbable luck.
\end{abstract}


\section{Introduction}
Can a computing machine learn to achieve all goals in a complex environment? How complex a machine needs to be in order to do so? In this paper we attempt to define and ask these questions and discuss a possible answer, and the constraints and limitations the answer implies for all learning machines or algorithms.

Machine learning is a widely studied area. In most cases machine learning deals with supervised or assisted learning where information about the environment is passed to a learning machine for training during the learning phase, for example, a list of labeled examples of patterns for learning a concept. Assisted or supervised learning is a mature field and good progress has been made over the years, for example, back propagation training for multilayer neural networks \cite{RHW86}, support vector machines \cite{VC95}, and probably approximately correct (PAC) model for computational learning \cite{Val84}.

Discovering which patterns are important for achieving a certain goal in an unknown environment is an important learning problem in its own right. Machine learning also studies unsupervised learning, for example, for neural networks to learn the statistical structure in input data without training examples \cite{HS99}, or for clustering of unlabeled data, or for feature extraction or dimensionality reduction for large input datasets, which are steps in this direction.

In this paper, we study unassisted learning in an unknown environment, where a learning machine does not receive any information about its environment from an external agent. Henceforth by learning we always mean unassisted learning in this sense.

\section{A Learning Agent Abstraction}

To achieve goals in an environment, a machine needs a body with sensors to sense and actuators to manipulate its environment. We use the term ``agent-body'' for this, and assume that the agent-body has sufficient sensing and physical action capabilities for achieving a sufficiently large set of goals in a complex environment to allow a meaningful discussion of goal achieving and learning capabilities. Defining exact specifications of the agent-body is not relevant to the thesis presented here except that when we compare the goal achieving capabilities of different machines, we assume that they all have physically identical agent-bodies. 

A machine also requires a computing platform part and an information processing algorithm to be able to achieve goals in a complex environment. We define the term  ``agent'' as the algorithm or the Turing machine (both used equivalently here) running on the computing platform part which processes all the incoming sensory information and controls all the physical actions of the agent-body. We consider two agents $A$ and $B$ identical only when a binary string\footnote{a bit string, i.e., a member of $\{0, 1\}^*$. The choice of alphabet is immaterial.} representation for $A$ including all its data is identical to that for agent $B$ under the same representation scheme. As an agent learns it may change itself by modifying its code and data. Let $start\_size(A)$ denote the size in number of bits for the binary string representation for an agent $A$ including all its data at the start of learning experiments. We assume an agent to have as much blank memory available for its use as it requires. It is important to emphasize here that in this paper we do not consider agents constrained by a limited amount of memory. When we use the expression ``finite sized agent" for an agent $A$, it means a finite $start\_size(A)$ with the assumption that the agent $A$ is allowed to use unlimited amount of memory and grow to any arbitrary size.  
An extension of the Church-Turing Thesis \cite{Tur48} is premised that states that the learning and goal achieving capabilities of any physically realizable system can be implemented by a Turing machine running on a computing platform of comparable throughput, and controlling an agent-body which is physically equal in terms of physical action capabilities and sensory capabilities. This is equivalently stated by the physical symbol system hypothesis of Newell and Simon \cite{NS76}:

   \textsl{``A physical symbol system has the necessary and sufficient means for general intelligent action.''}

We define a \emph{goal} to be a pair of environment states $(s_1,s_2)$, where $s_1$ is the starting state and $s_2$ is the end state. The physical state of the agent-body outside the computing platform part is considered to be a part of the environment state. We use the term \emph{goalset} to denote a set of goals. Let $ALL\_GOALS$ denote the set of all goals in an environment that the agent-body can physically achieve. 

An agent achieves a goal $ g = (s_1,s_2) $ using some sequence of actions that takes the environment through a path, that is, a sequence of states, which has cost, for example, energy cost, environmental cost, elapsed time cost, and possibly other costs. We represent the sum of all the costs for achieving a goal $g$ through some path by a positive real number and denote it by $cost(g)$. We note that $cost(g)$ can be different for different paths used for the same goal $g$. Different paths for the same goal $g$ can result from the agent not repeating the same sequence of actions, or from the randomness in the environment not resulting in the same state transitions for the same actions. We note that the amount of memory used or the number of computing steps executed by an agent in achieving a goal $g$ is not counted in the measurement of the $cost(g)$\footnote{although \emph{energy} and \emph{elapsed time} may be included in cost measurement, the thesis does not impose any limit on the number of computational steps that can be executed in any given elapsed time period or amount of energy. It will be an interesting result to show a dependence of the thesis presented here on a theoretically infinite computing speed. We note that in our universe there will be an upper limit for any physical computing device on how many computational steps can be executed within a given time period or amount of energy.}.

To determine if an agent knows how to achieve a goal $ g = (s_1,s_2) $, we set a ``target cost'' for each goal \nopagebreak[4] using the function\footnote{choice of pcost() is arbitrary and immaterial for the thesis presented here, and we can choose the value of $pcost(g)$ appropriately approaching the optimal cost for each $g$ to indicate the agent's goal achieving capability within a certain cost with any good enough success rate.} $ pcost:ALL\_GOALS \rightarrow \mathcal{R}^+ $, and say that an agent $A$ knows how to achieve a goal $g$, or use the phrase ``an agent $A$ can \emph{effectively} achieve a goal $g$'' if and only if the agent $A$ can achieve the goal $g$ with $ median(cost(g)) \leq pcost(g) $ in repeated trials with identical start state $s_1$ and identical agent $A$ at start. When we say ``an agent $A$ can \emph{effectively} achieve a goalset $G$'' we mean $ \forall g \in G$, \textsl{$A$ can effectively achieve $g$}.

By exploring its environment, an intelligent agent can learn to effectively achieve a goal $g$ even if it can not effectively achieve the goal $g$ to start with. Let us denote the cost for all the explorations by an agent $A$ for learning to effectively achieve a goal $ g = (s_1,s_2) $ starting from state $s_1$ by $ learning\_cost(g,A) $. This cost would depend on $g$, as well as on the agent $A$ and the state of its knowledge about its environment. Also, this cost may vary if we repeat the experiment to find $ learning\_cost(g,A) $ with identical start state $s_1$ and identical agent $A$. Let $ learning\_cost(G,A) $ denote the cost of learning to effectively achieve every goal in a goalset $G$ for an agent $A$.

To determine if an agent $A$ can \emph{efficiently}\footnote{the term ``efficient" is used here differently than the usage of the term to mean polynomial time deterministic computation in computational complexity theory.} learn to effectively achieve a goalset $G$, we set a ``target learning cost'' for each goalset using the function $ lcost:\mathcal{P}(ALL\_GOALS) \rightarrow \mathcal{R}^+ $, and use the phrase ``agent $A$ can \emph{efficiently} learn to effectively achieve goalset $G$'' or ``agent $A$ can \emph{efficiently} learn goalset $G$'' if and only if $ median(learning\_cost(G,A)) \leq lcost(G)$ in repeated trials with identical agent $A$ at start.

\section{Principles of Learning}

An agent needs information about its environment to be able to effectively achieve goals in a complex environment. We hypothesize that there is no finite body of information that an agent can start with and \emph{effectively} achieve any arbitrary goalset in an environment of unbounded complexity. The $start\_size$ of the smallest agent that can \emph{effectively} achieve a given goalset can be no smaller than a certain critical size, and the $start\_size$ would grow with larger (superset) goalsets requiring new environmental information. 

We define the term $Description\_Complexity(G)$ for a goalset $G$ in an environment $S$ as an integer value, such that for all agents $A_i$ that can \emph{effectively} achieve $G$, $start\_size(A_i) \geq Description\_Complexity(G)$\footnote{This is an extension of the idea of Kolmogorov complexity, also called Descriptive complexity, which is defined for strings.  $Description\_Complexity(G)$ extends the concept to define the complexity of a goalset $G$ in an environment $S$, assuming the agent-body to be a part of the environment $S$.}. The first principle is stated below.

\emph{Principle of Increasing Description Complexity.}\\
For every goalset $G_i \subsetneq ALL\_GOALS$, there is a goalset  $G_j \subseteq ALL\_GOALS$ such that 
$Description\_Complexity(G_i \cup G_j) > Description\_Complexity(G_i).$\\

The first principle effectively says that the $Description\_Complexity$, or the $start\_size$ of the smallest agent that can effectively achieve a goalset, grows monotonically with larger (superset) goalsets. The $Description\_Complexity(G)$  grows to infinity as $G$ grows to $ALL\_GOALS$ for an environment of unbounded complexity.

We will use the phrase ``$Description$ of environment $S$ with respect to a goalset $G$", or in short ``$Description$" when $S$ and $G$ are clear from the context, to denote an agent of $start\_size = Description\_Complexity(G)$ that can effectively achieve $G$. We note that the smallest agent for a goalset $G$ may not be unique. In other words, there can be multiple alternative $Descriptions$ of an environment $S$ with respect to a goalset $G$.

We discuss now the constraints for learning that all agents would be limited by.  
We hypothesize that there is no general, domain independent algorithm that would allow an agent to efficiently learn any arbitrary goalset in an environment starting with zero information about the goal domain. An agent must have environment representational building blocks, or what we will call $Microconcepts$ henceforth, to be able to efficiently learn a required $Description$ of its environment to be able to effectively achieve goals. $Microconcepts$ are bit-strings that are found by evolutionary search and bring an agent higher up in the fitness landscape to allow an agent to efficiently learn a goalset. The phrase ``efficiently learn'' as defined above is important here. An algorithm that learns by evolutionary change, starting with zero information about its environment,  can learn to achieve any arbitrary goalset, but only over a very large evolutionary timescale. Here the term evolutionary change means a change that is not a pre-calculated change made for a known fitness benefit. The fitness benefit of change towards the knowledge of the environment and the target goalset can only be evaluated by experimenting in the environment after the change is made, since the agent does not have the required knowledge of the environment for an evaluation beforehand. We also assume that the luck of any physical system is limited by the laws of probability. In the case of agents, this means that the probability of success of generating a large Description bit string by random guessing would be exponentially diminishing with the increasing size of the Description string. The second principle states this hypothesis below.

\emph{Principle of Microconcepts for Learning.}\\
For an environment $S$ and a goalset $G$ on $S$, an agent must have, at the start, a set of goalset specific bit strings, denoted $Microconcepts(G)$, from a class of possible such sets, to be able to efficiently learn the goalset $G$. For any agent $A$ and a goalset $G'$ for which $A$ does not have a complete required set for a possible $Microconcepts(G')$, $A$ would be limited to an ``inefficient'' learning through evolutionary change to learn the missing members of $Microconcepts(G')$ before it can have the capacity to efficiently learn $G'$.

Let us define $Critical\_Agent\_Size(G)$ for a goalset $G$ in an environment $S$ as an integer value, such that for all agents $A_i$ that can efficiently learn $G$, $start\_size(A_i) \geq Critical\_Agent\_Size(G)$. We extend the principle of increasing complexity to learning agents with the third principle stated below.

\emph{Principle of Increasing Learning Complexity.} \\
For any agents $A_1$ and $A_2$ that can efficiently learn goalsets $G_1$ and $G_2$ respectively and have sizes $Critical\_Agent\_Size(G_1)$ and $Critical\_Agent\_Size(G_2)$ respectively,

$G_2 \supsetneq G_1 \rightarrow Critical\_Agent\_Size(G_2) \geq Critical\_Agent\_Size(G_1).$\\

For all goalsets $G_1$ and $G_2$ such that $G_2 \supsetneq G_1$ and all possible $Microconcepts(G_2) \backslash Microconcepts(G_1)$ are non empty sets,
$Critical\_Agent\_Size(G_2) > Critical\_Agent\_Size(G_1)$.

That is, agents that can efficiently learn larger (superset) goalsets requiring new environmental information have larger $start\_size$.

\section{Consequences}

How much slower evolutionary learning is compared to efficient learning? Let us consider the cost of evolutionary learning for a goalset $G$ starting with zero knowledge of the environment. We have defined $lcost(G)$ as the cost of efficient learning. The $start\_size$ of the smallest agent that can efficiently learn the goalset $G$ is defined to be $Critical\_Agent\_Size(G)$. If we start with an agent $A$, and if $C$ is the number of bit differences, or the Hamming distance, between $A$ and the closest agent $A'$ that can efficiently learn the goalset $G$, the principles of learning imply that the only way to learn these $C$ bits is through evolutionary change. Since we start with an agent with zero information about the environment, $C = Critical\_Agent\_Size(G)$. Let us denote by $Evol\_Cost(G)$ the median cost of evolution of an agent that can efficiently learn a goalset $G$. 

The fitness function for evolution measures the capability of an agent to learn the goalset $G$. The fitness landscape is over all possible agent codes, that is, all possible sequences of $C$ bits which gives it a size of $2^C$ agent code points. Apart from the size, the other complexity of the fitness landscape is the bit distances, i.e., the Hamming distances, to be searched from any code point in the fitness landscape to move to a point of higher fitness value. We assume a fitness landscape such that from any code point\footnote{except for the final evolution point that corresponds to an agent that can efficiently learn $G$.}, there is a point of higher fitness value within a Hamming distance of $R$ bits.

Since we are starting with zero knowledge of the environment, the cost of evaluation of the fitness function will be based on testing an agent's learning capability through the agent's learning exercise in the environment. We assume that the cost of evaluation of the fitness function will grow linearly with the number of bits the agent accumulates towards the right sequence of $C$ bits as the agent moves higher in the fitness landscape and the agent's capability grows to efficiently learn larger subsets of $G$. Since we are interested in estimating how fast evolutionary learning is (or how slow evolutionary learning is compared to efficient learning), we will use a steepest ascent hill climbing algorithm to estimate the growth of $Evol\_Cost(G)$ with the growth of $R$ and $C$.

Let $\kappa$ denote the proportionality constant that relates the cost of the fitness function to evaluate if an agent can efficiently learn a goalset $G$ to the smallest size of the agent that can efficiently learn the goalset $G$. As an agent evolves to have the capability to efficiently learn larger (superset) goalsets leading up to $G$, $R$ new bits are added in each generation, and the number of added bits grows to $C$ bits in $C/R$ generations. The cost of evaluation for the $i$th generation is $2^R \times \kappa \times R \times i$, since all $2^R$ agents are evaluated to select the agent for creating the next generation, and $R \times i$ is the total size of the accumulated bits in the agent being evaluated in the $i$th generation. For the median cost of evolution we sum the cost of evaluation of each generation of agents except the final generation for which we count half the cost.
Therefore, 
\begin{equation}
Evol\_Cost(G) = \sum_{i=1}^{C/R} 2^R \times \kappa \times R \times i -  \frac{1}{2} \times 2^R \times \kappa \times R \times \frac{C}{R}
\end{equation}

Or, 
\begin{equation}
Evol\_Cost(G) = \frac{1}{2} \times 2^R \times \kappa \times \frac{C^2}{R}
\end{equation}

The evolution cost for an agent that can efficiently learn a goalset $G$ starting with zero knowledge of the environment grows exponentially with $R$ and grows as the square of the size $C$ of the smallest agent that can efficiently learn $G$. 


If we take $lcost(G)$ as the fitness function evaluation cost for an agent's efficient learning capability for a goalset $G$, and $C$ is the size of the smallest agent that can efficiently learn $G$, we have $lcost(G) = \kappa \times C$. Therefore, 

\begin{equation}
Evol\_Cost(G) = \frac{1}{2} \times 2^R \times \frac{C}{R} \times lcost(G) 
\end{equation}

The thesis presented here implies that the evolutionary learning cost for a goalset $G$ for an agent which does not have the required $C$ bits at the start is of the order of $(2^R \times \frac{C}{R})$ times larger than the efficient learning cost for an agent that has the required set of $Microconcepts(G)$ and a $start\_size$ of $C=Critical\_Agent\_Size(G)$ bits. 


We can estimate the value of $\kappa$ given by 
\begin{equation}
\kappa = \frac{lcost(G)}{Critical\_Agent\_Size(G)}
\end{equation}

if we have reasonable estimates for the values of $lcost(G)$ and $Critical\_Agent\_Size(G)$ for a goalset $G$. Here we estimate the value of $\kappa$ by assuming the size of the smallest agent in an earth like environment that can learn in 25 years the goalset that an average human can learn in 25 years to be 400 million bits\footnote{The total size of useful information in the human genome is variously estimated to be in the range of 200 to 800 million bits. The size of the smallest algorithm that has the same learning capability as an average human could be much different than the size of the useful information in the human genome.}. If we measure the evaluation cost in ``year" units\footnote{where 1 ``year" equals the total cost of one person-year of efforts by a human equivalent agent in an earth like environment.}, then the evaluation cost for evaluating the learning capability of such an agent would be 25 years. The value of $\kappa$ thus can be estimated as
$$
\kappa = \frac{25\ years}{400,000,000\ bits} \approx 0.6 \times 10^{-7}\ year/bit
$$

Using this value of $\kappa$ on an earth like environment, for an example, we estimate the $Evol\_Cost(G)$ for an agent that can learn a goalset $G$ in 25 years that an average human can learn in 25 years, with the following other assumptions: 1. the environment can support one billion simultaneous agents in one generation, 2. $C = 400\ million\ bits$, and 3. $R=30\ bits$ for this example, to make $2^R$ equal to the number of agents in one generation so that we can get out of a local maxima in one generation of 1 billion agents. We find the evolution cost using (2) as 
$$
Evol\_Cost(G) \approx \frac{1}{2} \times 2^{30} \times 0.6 \times 10^{-7} \times \frac{(400 \times 10^6)^2}{30}\ years
$$
Or,
$$
Evol\_Cost(G) \approx 1.7 \times 10^{17}\ years
$$

We compute the elapsed time for this evolution by dividing the above cost by $10^9$, the number of agents in one generation:
$$
Elapsed\ time \approx  \frac{1.7 \times 10^{17} }{10^9} = 170 \ million\ years
$$

An agent that has, at the start, the required $Microconcepts$ for the goalset $G$ with a total $start\_size$ of 400 million bits (in this example) can learn the goalset $G$ efficiently in 25 years, while another agent which is starting with zero environmental information to start with, can only learn the goalset $G$ inefficiently using evolutionary methods in around $1.7 \times 10^{17}$ years using the evolutionary algorithm used in this example, which possibly can be improved upon, but not enough to make its cost comparable to the efficient learning cost.

\section{Conclusion}

We have presented the thesis that more capable agents have larger complexity, or more precisely, the smallest agents that can efficiently learn larger goalsets in a complex environment have larger $start\_size$.

We have proposed that for efficient learning, a machine must start with some pre-encoded information about its environment, or $Microconcepts$, which can only be gained through evolutionary changes, and which decide the efficient learning boundaries for a machine. A machine can only learn beyond its efficient learning boundaries at a much slower pace using evolutionary methods. This also discounts the possibility where we would cross a certain threshold in AI development which would allow an AI system to recursively create increasingly more powerful AI systems in a very short span of time, thereby creating a system of much larger intelligence.

\subsection{Future work}
In this paper we have made simplifying assumptions for estimating the cost of evolution, for example, asexual evolution. If we allow all possible methods, what is the lowest cost of evolution of an agent that can efficiently learn a given goalset $G$ in an environment $S$, starting with an agent with zero information about the environment $S$? 

Are the principles presented here derivable as a consequence of $P \neq NP$ (assuming $P \neq NP$) \cite{Aar11,Wig07}?

\section*{Acknowledgement}
I thank Scott Aaronson for reviewing an earlier draft and providing valuable suggestions. I would also like to thank the  web for providing easy access to the vast amount of work done in this field and to all the people (too numerous to mention here) on whose work this work builds on.


\nocite{*}
\bibliographystyle{IEEEtran}

\bibliography{IEEEabrv,paper_evolution_ieee.bib}

\end{document}